\documentclass[11pt,a4paper]{article}
\usepackage[hyperref]{acl2021}
\usepackage[noabbrev,capitalise]{cleveref}
\usepackage{times}
\usepackage{latexsym}
\usepackage{graphicx}
\usepackage{booktabs}
\usepackage{amssymb}
\usepackage{enumitem}
\usepackage{soul}
\graphicspath{ {.} }

\usepackage{microtype}

\aclfinalcopy 

\title{Unsupervised Topic Segmentation of Meetings with BERT Embeddings}

 \author{Alessandro Solbiati \\
  Facebook, Inc. \\
  \texttt{lessandro@fb.com} \\
  \textbf{Shivani Poddar}\\
  Facebook, Inc. \\
  \texttt{shivanipi@fb.com} \\\And
  Kevin Heffernan \\
  Facebook, Inc. \\
  \texttt{kheffernan@fb.com} \\
  \textbf{Shubham Modi} \\
  Facebook, Inc. \\
  \texttt{shubhammodi@fb.com} \\\And
  Georgios Damaskinos \\
  Facebook, Inc. \\
  \texttt{damaskinos@fb.com} \\
  \textbf{Jacques Cali} \\
  Facebook, Inc. \\
  \texttt{jcali@fb.com} \\
}
\date{}

\begin{document}

\maketitle
\begin{abstract}
Topic segmentation of meetings is the task of dividing multi-person meeting transcripts into topic blocks. Supervised approaches to the problem have proven intractable due to the difficulties in collecting and accurately annotating large datasets. In this paper we show how previous  unsupervised topic segmentation methods can be improved using pre-trained neural architectures. We introduce an unsupervised approach based on BERT embeddings that achieves a 15.5\% reduction in error rate over existing unsupervised approaches applied to two popular datasets for meeting transcripts.
\end{abstract}

\section{Introduction}

With remote work being the norm, the average employee attends 62 company meetings every month \citep{meetings}.
The majority of existing video call tools used for professional meetings, enable a recording functionality that is either rarely used or its output is stored and never used again after the meeting.
Nevertheless, these meeting recordings create a wonderful opportunity for increased productivity and transparency.

Topic segmentation is the task of dividing text into a linear sequence of topically-coherent segments. 
In the context of meeting recordings and their transcripts, topic segmentation can quickly provide users with a valuable high level understanding of past meetings.
For example, upper management can quickly locate critical decisions taken during a product meeting among engineers. Topic segmentation can also significantly boost search indexing and downstream retrieval~\cite{hearst2002}, where keyword search is not effective given the usually high transcription error rate of Automatic Speech Recognition (ASR) systems. 
Finally topic segmentation can be useful for other text analysis tasks such as passage retrieval~\cite{salton96}, document summarization and discourse analysis~\cite{galley2003discourse}.

However, topic segmentation of meetings is a very challenging task due to (a) the noisy nature of meeting transcripts and (b) the lack of ground truth data.
First, these meetings involve multiple participants with each having their own personalised use of the language thus inevitably leading to transcript errors.
Second, topic segmentation can be a hard task even for human annotators~\cite{Gruenstein2008}, and hence collecting labeled data for segmented meetings becomes complex and expensive. 
In addition, organisations express strong sensitivity towards their private meeting data, making the task of collecting large training datasets even harder. 

In this paper we focus on \emph{unsupervised} topic segmentation of meetings.
The lack of ground truth data impedes any benefits from the latest advances in neural networks \citep{barrow2020}, as opposed to segmentation in other domains such as written text where a large amount of labeled data recently brought significant advancements through supervised neural models \citep{badjatiya2018attention}.

We remove this impediment by proposing a mechanism based on pre-trained transformer models~\cite{vaswani2017attention} applied to the task of topic segmentation of meetings.  
Our method is completely unsupervised, i.e., does not require any training data.
Our model utilizes a new similarity score based on BERT embeddings \citep{devlin2018bert}, that enables a 15.5\% reduction in error rate compared to existing unsupervised methods that use similarity score heuristics not based on neural models.
We also show a 26.6\% reduction in error rate compared to the current state of the art supervised topic segmentation models \citep{badjatiya2018attention} trained on text datasets.
These models perform poorly due to the distinctive differences between written text datasets such as Wikipedia, and the standard meeting transcripts datasets ICSI Meeting Corpus~\cite{janin2003icsi} and AMI Meeting Corpus~\cite{kraaij2005ami}.

\section{Related Work}

\paragraph{Written text.} There are many recent advancements on topic segmentation of written text, with most of them
being based on bidirectional-LSTM embeddings.
\cite{li2018segbot} combine a BiLSTM with a pointer network,
\citep{badjatiya2018attention} propose stacked BiLSTMs with attention for topic segmentation of an 85 fiction books dataset~\cite{kazantseva2011linear}, and
\citep{barrow2020} propose a custom LSTM architecture for  topic segmentation on the WikiSection dataset~\cite{wikisection} that consists of 242k labeled segments from Wikipedia articles.

\paragraph{Monologue transcripts.} Topic segmentation of spoken language is significantly more challenging than written text due to the added complexity that the underlying ASR system introduces. 
This task focuses either on monologue data or on dialogue data~\cite{purver2011topic}. 
Monologue data also witnessed recent advancements with neural-based architectures such as TCNs~\cite{zhang2019topic} and Bi-LSTMs~\cite{sehikh2017topic}. 
Monologue datasets also feature an abundance of labeled training data, mainly comprised of large broadcast news transcripts such as the Euronews Dataset with 24k labeled segments~\cite{sheikh2016diachronic}.

\paragraph{Multi-party dialogue transcripts.} Multi-party dialogue speech data, mainly comprised of meeting transcripts, has not yet benefited from the advancements that other sub-domains have seen with neural networks.
Most of the existing methods in this domain are based on measuring similarity/coherence between sentences to detect topic changes. One of the first successful approaches of this sort has been TextTiling~\cite{hearst1997text}, that uses word frequency vectors as a similarity metric. 
Despite TextTiling being originally designed for text documents, it has been successfully applied to meeting transcripts segmentation~\cite{georgescul2006analysis}. 


\paragraph{Sentence embeddings.}
These have been used to extract semantic similarity, a task formalised in~\cite{cer2017semeval} as the degree to which two sentences are semantically equivalent.
BERT~\cite{devlin2018bert} is a pre-trained transformer network~\cite{vaswani2017attention} which reaches state-of-the-art-results 
for many NLP tasks. 
No independent sentence embeddings are directly computed in the original model, hence a common practice is to derive a fixed vector by either averaging the outputs or by using the outputs special CLS toke~\cite{zhang2019bertscore}.
Sentence-BERT~\cite{reimers2019sentence} is a modification of BERT designed to derive semantically meaningful sentence embeddings. 
Sentence-BERT employs siamese and triplet network structures~\cite{schroff2015facenet} to derive embeddings that can be compared with cosine similarity.


\begin{table*}
\centering
\scalebox{0.8}{
\begin{tabular}{lcp{8.5cm}c}
\toprule
\textbf{Topic Label} & \textbf{Topic Change} & \textbf{Caption} & \textbf{Speaker}\\
\toprule
What to do for next meeting & 0 & Yeah, since they’re not at the meeting I think it’s in [disfluency] out of courtesy we should first ask them. & C \\
. & 0 & Yes. & D \\
. & 0 & And I'll try to [disfluency]  & A \\
. & 0 & Yes. & D \\
Coffee Availability & 1 &  Bu I ju just before finishing uh, I mean, we have a cafeteria or we don't eat at all?  & B \\
. & 0 & Fine. & A \\
. & 0 & We don’t have a cafeteria. & D \\
. & 0 & What do you mean by cafeteria? & A \\
\bottomrule
\end{tabular}
}
\caption{Example of meeting transcript from AMI Meeting Corpus (ID IB4003)  as a sequence labelling problem.}
\vspace{-4mm}
\label{table:transcript}
\end{table*}

\section{Method} \label{sec:method}

In this section we provide a formal presentation of the topic segmentation task, and a detailed overview of our model.

\emph{Input}: a meeting transcript produced by an ASR system consists of a list of $M$ utterances $S = \{S_1, \cdots, S_M\}$ with an underlying topic structure represented by a reference topic segmentation $T = \{T_1, \cdots, T_N\}$, with each topic having a start and an end utterance $T_i \in [S_j, S_k]$.

\emph{Output}: a label sequence $Y = \{y_1, \cdots, y_M\}$ where $y_i$ is a binary value that indicates whether the utterance $S_i$ is the start of a new topic segment.

Our topic segmentation model consists of (i) a sentence representation model to extract semantic similarity between sentences and (ii) a segmentation scheme that employs semantic similarity variations over time to detect topic changes.

\subsection{Sentence Representation}
To extract semantic similarity we experiment with two different sentence representation approaches.
\paragraph{BERT.} The first pre-trained model we use is RoBERTa~\cite{liu2019roberta}, a pre-training configuration of BERT trained with the Masked Language Modelling objective on a five large English-language corpora totaling over 160GB of uncompressed text \citep{zhu2015aligning}. 
It is possible to extract fixed features from the pre-trained model without additional fine-tuning~\cite{devlin2018bert}: we extract a fixed sized vector via max pooling of the second to last layer. RoBERTa is trained with hyperparameters $L=12$ and $H=768$, where $L$ is the number of layers and $H$ is the size of the hidden layer. A sentence of $N$ words will hence result in an $N*H$ embedding vector. The closer to the last layer, the more the semantic information carried by the weights~\cite{zeiler2011adaptive}; hence our choice of the second to last layer. 

\paragraph{Sentence-BERT.} The second pre-trained model we experiment with, is the current state-of-the-art in sentence representation, namely Sentence-BERT~\cite{reimers2019sentence}, pre-trained on the SNLI dataset~\cite{bowman2015large}. We extract fixed size sentence embeddings using a mean over all the output vectors, similar to the method we used for BERT.

\paragraph{Max pooling.} Our extraction architecture is particularly robust to noisy speech data~\cite{shriberg2005spontaneous}, including ASR miss-transcriptions, disfluencies of speakers or turn-taking. 
A sample of this noisy characteristics can be seen in \cref{table:transcript} where we report the transcript of 8 example utterances. To filter out words that hold limited semantic value (e.g., ``uh, I mean.''), we apply repeatedly a max pooling operation to extract words with high semantic value from a given utterance.

\subsection{Segmentation Scheme}
Given a valid sentence embedding, a common practice is to train a supervised classifier to perform sequence labelling, for example using TCNs~\cite{zhang2019topic}. 
On the contrary, we choose a completely unsupervised approach that does not require any labeled training data. 

\ Our approach is a modified version of the original TextTiling~\cite{hearst1997text} ``gold-standard unsupervised method for topic segmentation''~\cite{purver2011topic}. TextTiling detects topic changes with a similarity score based on word frequencies, whereas \emph{we detect topic changes based on a new similarity score using BERT embeddings} as follows.
\begin{enumerate}[topsep=0pt,itemsep=-1ex,partopsep=1ex,parsep=1ex,leftmargin=0.3in,align=left]
    \item Compute BERT embeddings for every utterance $S_i$ of the meeting transcript.
    \item Divide the meeting corpus into blocks of utterances $\{S_i, \cdots, S_k\}$, and perform a block-wise max pooling operation to extract the embedding $R_i$ for each block. 
    \item Compute cosine similarity $sim_i$ between adjacent blocks $R_i$ and $R_{i+1}$, where $sim_i$ represents the semantic similarity between two blocks separated at utterance $S_i$.
    \item Derive the topic boundaries as pairs of blocks $R_i$ and $R_{i+1}$ with semantic similarity $sim_i$ lower than a certain threshold.
    In particular, we obtain a sequence of topic changes ${T = \{i \in [0, M] | sim_i < \mu_s - \sigma_s \}}$ where $\mu_s$ and $\sigma_s$ are the mean and variance of the sequence of block similarities $sim_i$.
\end{enumerate}

\section{Evaluation} \label{sec:eval}
\paragraph{Datasets.}
To demonstrate the effectiveness of our model we evaluate it on the two major collections of meeting data produced in recent years. The \emph{ICSI Meeting Corpus}~\cite{janin2003icsi} includes 75 recorded and transcribed meetings with topic segmentation annotations~\cite{gruenstein2008meeting} and the \emph{AMI Meeting Corpus}~\cite{kraaij2005ami} includes 100 hours of recorded and transcribed meetings also with  topic segmentation annotation. Both datasets include a hierarchical structure of the topic annotation.
For the purpose of this paper we  consider only the top-level meeting changes, i.e., we perform linear topic segmentation. 

Despite the fact that AMI and ICSI annotations could be used to train a small supervised segmenter model, in a practical application of meeting segmentation there will be often no labeled data available given the complexity of the annotation task. Hence our unsupervised evaluation methodology is representative of real world practical meeting segmentation scenarios.

\begin{figure}[t]
    \centering
    \includegraphics[scale=0.3]{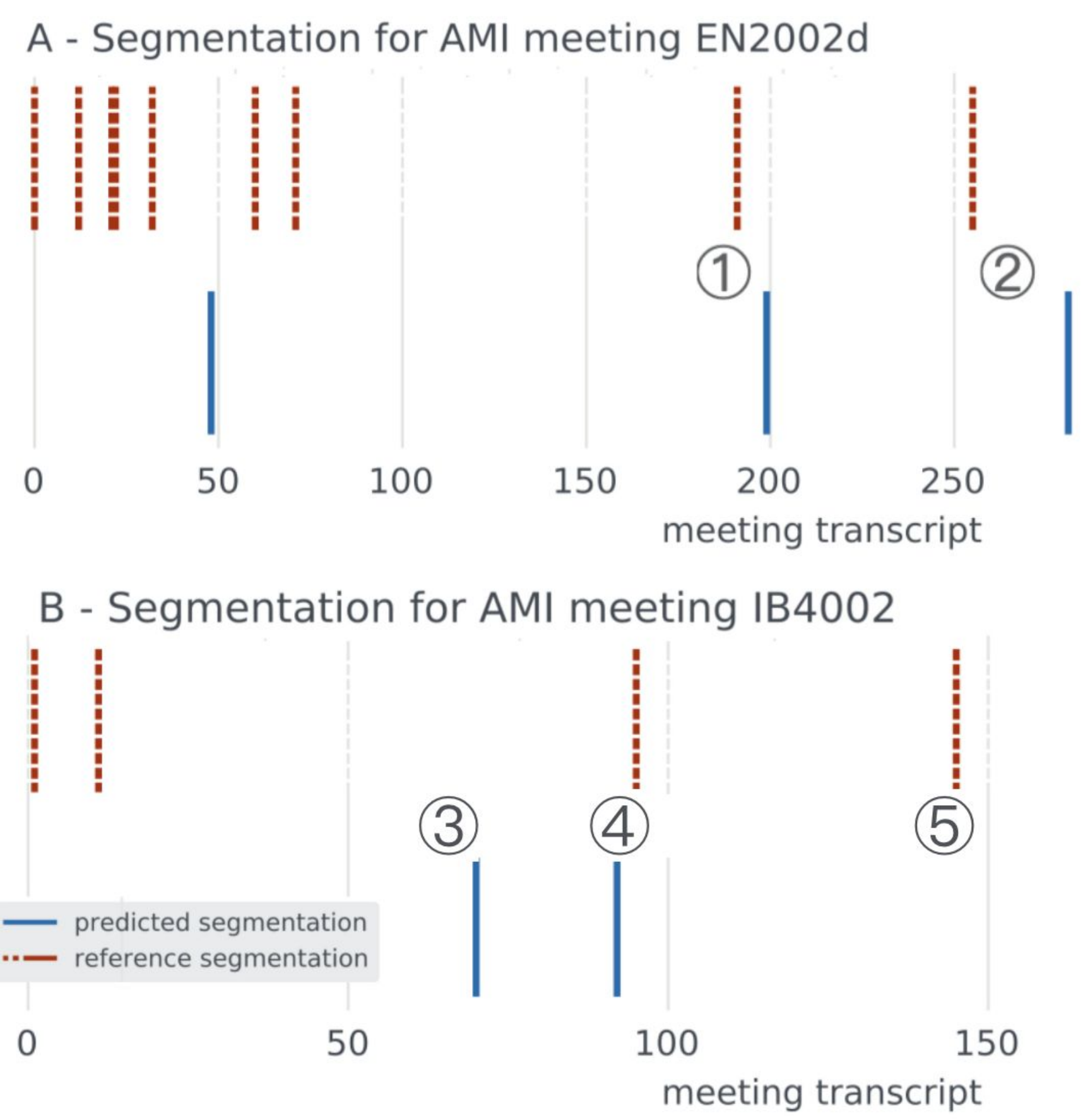}
    \caption{Comparison of predicted segmentation by our BERT model (in blue) and reference segmentation by human annotators (in dotted brown) on two meetings of the AMI dataset. Meeting EN2002d is an example of a  low error rate segmentation (\emph{Pk/WinDiff}) as topic changes \raisebox{.5pt}{\textcircled{\raisebox{-.9pt} {1}}} and \raisebox{.5pt}{\textcircled{\raisebox{-.9pt} {2}}} are \emph{true positives}. Meeting IB4002 is an example of high error rate segmentation as \raisebox{.5pt}{\textcircled{\raisebox{-.9pt} {3}}} is a \emph{false positive} topic change, \raisebox{.5pt}{\textcircled{\raisebox{-.9pt} {5}}} is a \emph{false negative} topic change and only \raisebox{.5pt}{\textcircled{\raisebox{-.9pt} {4}}} is a \emph{true positive} topic change.}
    \label{fig:my_label}
    \vspace{-4mm}
\end{figure}

\paragraph{Metrics.}
To evaluate the performance of our model we use two standard evaluation metrics, namely \emph{Pk}~\cite{beeferman1999statistical} and \emph{WinDiff} ~\cite{pevzner2002critique}. 
Both metrics use a fixed sliding window over the document, and compare the predicted segmentation with the reference segmentation from the annotations to express a \emph{probability of segmentation error}. \cref{fig:my_label} depicts two predicted segmentations by our model; one with a high and one with a low Pk/WinDiff score.

\paragraph{Baselines.} First, we compare our method against two commonly referenced naive baselines as in \citep{beeferman1999statistical}. The \emph{Random} method places topic boundaries uniformly at random and the \emph{Even} method places boundaries every n-th utterance. 
Second, we compare our method against state-of-the-art supervised learning models for topic segmentation \citep{badjatiya2018attention}. 
These models are based on BiLSTM-CNN architectures \citep{hochreiter1997long} that require large datasets to converge.
Hence the only viable approach is to train them on written text segmentation datasets \citep{kazantseva2011linear} such as fiction books or Wikipedia articles. Unfortunately, meeting transcripts are fundamentally different from written text, with most utterances conveying minimal semantic significance (\cref{table:transcript}).
Finally, we compare against the standard unsupervised topic segmentation baseline, namely TextTiling~\cite{hearst1997text}, that uses word frequencies to measure the similarity among the utterances.

\begin{table}
\centering
\scalebox{0.75}{
\begin{tabular}{lcccc}
\toprule
\textbf{Method} & \textbf{AMI Pk} & \textbf{AMI Wd} & \textbf{ICSI Pk} & \textbf{ICSI Wd}\\
\toprule
Random & 0.604 & 0.751 & 0.632 & 0.837 \\
Even & 0.513 & 0.543 & 0.601 & 0.660 \\
TextTiling &  0.391 & 0.410 & 0.382 & 0.408 \\
BiLSTM & 0.447 & 0.473 & 0.410 & 0.430 \\
\textbf{BERT (our)} & \textbf{0.331} & \textbf{0.333} & 0.337 & \textbf{0.345} \\
\textbf{S-BERT (our)} & 0.339 & 0.334 & \textbf{0.336} & 0.349 \\
\bottomrule
\end{tabular}
}
\caption{Performance of different topic segmentation methods using Pk and Wd (WinDiff) on AMI and ICSI datasets. \emph{Random} and \emph{Even} are the naive baselines; \emph{TextTiling} is the unsupervised segmentation baseline; \emph{BiLSTM} is the CNN-based supervised segmentation baseline trained on Wikipedia data; embeddings based on \emph{BERT} and \emph{Sentence-BERT} are our proposal.}
\label{table:results}
\vspace{-4mm}
\end{table}

\paragraph{Results.}
As shown in \cref{table:results}, the standard TextTiling method obtains the lowest error rate ($P_K = 0.382$) on ICSI dataset, in accordance to existing reports~\cite{georgescul2006analysis}.
Our methods based on BERT and Sentence-BERT embeddings obtain a 0.337 score on the same dataset, performing \emph{15.5\% and 11.8\% better} than TextTiling on the WinDiff and Pk metrics respectively for the ICSI Dataset.
The performance difference is justified as the BERT embeddings carry more semantic meaning compared to the word frequency scores of TextTiling. 
This richer semantic meaning allows for better detection of more nuanced topic changes.
Our BERT and Sentence-BERT embeddings have comparable performance on the tested datasets. 

The supervised BiLSTM model shows poor performance with a lowest error rate of $P_K = 0.410$ on ICSI dataset, compared to our best performance of $P_K = 0.337$. The supervised model is trained to segment fully formed sentences that carry substantially higher semantic signal compared to meeting transcript utterances. The performance difference is accentuated by the max pooling operation that makes our unsupervised method particularly robust to the noisy meeting transcripts.

\section{Conclusion}
We presented an unsupervised model based on BERT embeddings for segmenting meeting transcripts. 
Our new model leverages the strong semantic representation power of BERT alongside a new semantic similarity scoring technique, to enable unsupervised topic segmentation for meeting transcripts. 
Our model shows improved segmentation performance compared to the non neural-based approach, namely TextTiling. 
As part of our future work, we would like to incorporate additional signal to the BERT embeddings, such as speaker information, meeting agendas and cross-meeting features, in order to boost similar tasks such as meeting summarization~\cite{zhu2020hierarchical}.

\bibliographystyle{acl_natbib}
\bibliography{topic_segmentation}

\clearpage
\appendix
\section{Implementation Details}
\label{sec:appendix}
For a clear description of our proposed model and algorithm we refer to \cref{sec:method} of our main paper.
The experimental code used to implement our method and baselines can be found at \url{https://github.com/gdamaskinos/unsupervised_topic_segmentation}.

\paragraph{Pretrained models.}
The publicly available pretrained models used in our experiments are available for download by using the following links:
\begin{enumerate}
    \item{\textbf{RoBERTA}: \emph{roberta-base} from the huggingface transformers Python library at \url{https://huggingface.co/roberta-base}}
    \item{\textbf{Sentence-BERT}: \emph{stsb-roberta-base} from the Sentence-BERT Python library at \url{https://www.sbert.net/docs/pretrained\_models.html}} 
\end{enumerate}
\paragraph{Evaluation runtime.}
Our approach is unsupervised and there are no particular computational requirements.
We report here the prediction runtime on the entire ICSI and AMI dataset:
\begin{enumerate}
    \item \textbf{BERT embeddings method}: max run time 14 minutes, 33 seconds, CPU time 1 hour, 9 minutes, 57 seconds, AWS equivalent \$0.29
    \item \textbf{Naive and TextTiling baselines}: min run time 2 minutes, 26 seconds, max run time 4 minutes, 52 seconds
\end{enumerate}
\paragraph{Evaluation metrics.}
We use the standard evaluation metrics Pk and WinDiff, that are commonly used to represent error rates of segmenter systems.
We refer to \cref{sec:eval} for more details.
We are using in our methods the \emph{Natural Langauge Toolkit Library} (\url{https://www.nltk.org/}) reference implementation of the evaluation metrics:
\begin{enumerate}
    \item \textbf{Pk} is computed using \emph{nltk.metrics.segmentation.pk}
    \item \textbf{WinDiff} is computed using \emph{nltk.metrics.segmentation.windowdiff}
\end{enumerate}

\section{Evaluation Data}
We evaluate our methods and baselines on the two standard meeting corpora AMI and ICSI.
We refer to \cref{sec:eval} for more deatils. 
\paragraph{Dataset description.}
The datasets are publicly available and need to be consumed through the NTX meeting visualisation tool (\url{http://groups.inf.ed.ac.uk/nxt/index.shtml}). The datasets are used entirely as a test dataset, without a validation dataset given that our unsupervised approach relies purely on embeddings from pre-trained models. In the attached code we did not include the datasets since they are publicly available by using the links below.
\begin{enumerate}
    \item \textbf{AMI Corpus} consists of 100 hours of recorded meetings (\url{http://groups.inf.ed.ac.uk/ami/download/}) . Data are annotated by human annotators using the NXT tool following this annotation guidelines (\url{http://groups.inf.ed.ac.uk/ami/corpus/Guidelines})
    \item \textbf{ICSI Corpus} consists of 70 hours of recorded meetings (\url{http://groups.inf.ed.ac.uk/ami/icsi/download}) and follows the same human annotation process as AMI.

\end{enumerate}
\paragraph{Dataset preprocessing.}
In our code we apply some lightweight text preprocessing that can be found in \emph{dataset.preprocessing}. The preprocessing includes:
\begin{enumerate}
    \item removing filler words and lower casing
    \item filtering captions shorter than 20 characters since they would not result in good quality embeddings
\end{enumerate}

\end{document}